\documentclass[a4paper, 10pt, conference]{ieeeconf}      
\usepackage{FG2024}
\usepackage{times}
\usepackage{epsfig}
\usepackage{graphicx}
\usepackage{amsmath}
\usepackage{amssymb}
\usepackage{multirow}
\usepackage{xcolor}
\usepackage{booktabs}
\usepackage{gensymb}
\usepackage{amsfonts}
\FGfinalcopy 

\IEEEoverridecommandlockouts                              
\def\FGPaperID{****} 

\title{\LARGE \bf
CrossGaze: A Strong Method for 3D Gaze Estimation in the Wild
}

\author{\parbox{16cm}{\centering
   {\large Andy Cătrună, Adrian Cosma, Emilian Rădoi}\\
    {\tt\small andy\_eduard.catruna@upb.ro, ioan\_adrian.cosma@upb.ro, emilian.radoi@upb.ro}\\
   {\normalsize
   Faculty of Automatic Control and Computer Science, POLITEHNICA Bucharest}}
}

\begin{document}

\ifFGfinal
\thispagestyle{empty}
\pagestyle{empty}
\else
\author{Anonymous FG2024 submission\\ Paper ID \FGPaperID \\}
\pagestyle{plain}
\fi

\maketitle

\begin{abstract}
Gaze estimation, the task of predicting where an individual is looking, is a critical task with direct applications in areas such as human-computer interaction and virtual reality. Estimating the direction of looking in unconstrained environments is difficult, due to the many factors that can obscure the face and eye regions. In this work we propose CrossGaze, a strong baseline for gaze estimation, that leverages recent developments in computer vision architectures and attention-based modules. Unlike previous approaches, our method does not require a specialized architecture, utilizing already established models that we integrate in our architecture and adapt for the task of 3D gaze estimation. This approach allows for seamless updates to the architecture as any module can be replaced with more powerful feature extractors. On the Gaze360 benchmark, our model surpasses several state-of-the-art methods, achieving a mean angular error of 9.94\degree. Our proposed model serves as a strong foundation for future research and development in gaze estimation, paving the way for practical and accurate gaze prediction in real-world scenarios.
\end{abstract}

\section{Introduction}

\begin{figure*}[hbt!]
\centering
\includegraphics[width=0.85\textwidth]{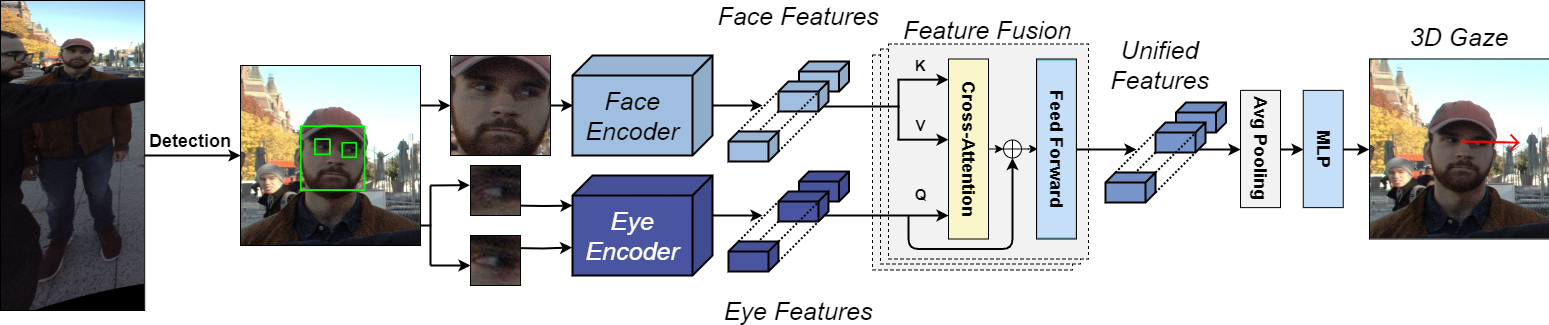}
  \caption{A high-level overview of the CrossGaze architecture. After the face is detected with a pretrained model, we process the features using a separate encoder for the face and for the eyes. The two resulting feature maps are processed using a cross-attention module to obtain the final gaze prediction.}
  \label{fig:architecture}
\end{figure*}

Human gaze estimation, the ability to infer the direction of a person's gaze from visual cues, is a critical aspect in understanding a person's intent, engagement and attention, which is highly valuable in fields such as automotive safety \cite{s22103959}, human-robot interaction \cite{10.3389/frobt.2022.770165} and virtual reality \cite{7892352}. Gaze estimation scenarios usually fall into two broad categories \cite{gaze-survey}: model-based and appearance-based. Model-based approaches employ specialized hardware and sensors such as near-infrared cameras (NIR) and are performed in constrained environments. Appearance-based methods, on the other hand, do not require specialized hardware and are meant to be used in unconstrained, real-world scenarios. While gaze estimation has been well-studied in controlled laboratory settings, where participants' head and eye movements can be tightly controlled, gaze estimation in the wild presents unique challenges due to the uncontrolled and dynamic nature of real-world environments. In the wild, individuals exhibit a wide range of head poses, lighting conditions, occlusions and distractions, making gaze estimation a challenging problem to solve. For commercial applications, gaze estimation in unconstrained environments can be essential for gathering actionable insights into customer behaviour \cite{eyeshopper}. Coupled with other uninstrusive soft-biometrics such as face and gait analysis \cite{catruna2021face,cosma22gaitformer}, gaze estimation from existing CCTV infrastructure enables a wide range of analytics which can be used to optimize customer experience and satisfaction.



In this work we present a series of simple improvements to the gaze estimation pipeline, showcasing state-of-the-art results without constructing a specialized architecture for this task. We combine the recent advancements in computer vision and image processing fields and construct a model which outperforms previous works on the Gaze360 benchmark. 

We propose the CrossGaze architecture which utilizes two separate encoders, one for the face and one for the eye region. This separation allows the face encoder to focus on extracting the global information as the local eye information is obtained by the second backbone. Our model leverages the cross-attention mechanism to combine the extracted features, which enables a prediction focused on the eye region that also accounts for the global information of the whole face.

This work makes the following contributions:

\begin{itemize}
    \item Our proposed method, CrossGaze, achieves a mean angular error of 9.94\degree\ on the Front 180\degree\ and 7.17\degree\ on the Front Facing subsets of the Gaze360 benchmark, surpassing several state-of-the-art methods. For CrossGaze we chose the most performant components according to our experiments, resulting in a strong baseline for gaze estimation in the wild.

    \item We provide an ablation study on the face encoder backbone, the procedure for incorporating eye features, and pretraining datasets. 
    Our results demonstrate that gaze estimation can be considerably improved by employing a multi-scale feature extractor, by pretraining on large-scale face datasets, and by incorporating eye-specific information through cross-attention.
\end{itemize}

\section{Related Work}

Recent years have seen a surge of interest in gaze estimation from images \cite{kellnhofer2019gaze360,zhang2017s,chen2018appearance,cheng2020coarse,abdelrahman2022l2cs}, driven by advances in computer vision architectures \cite{he2016deep,tan2019efficientnet,liu2022convnet,liu2021swin}. One of the main challenges in this task is training models that are sufficiently robust to estimate human gaze in unconstrained scenarios. Subsequently, progress in this direction is driven by the development of large and diverse datasets for gaze estimation in the wild \cite{zhang2015appearance,kellnhofer2019gaze360}, that capture subjects in natural settings. Zhang et al. \cite{zhang2015appearance} introduced the MPIIGaze dataset, which contains 200K images of 15 subjects in various settings. MPIIGaze has some limitations, such as the release of eye patches, and not full face images, or the small amount of participants which can lead to overfitting. 

Kellnhofer et al. \cite{kellnhofer2019gaze360} tackled some of the limitations of the MPIIGaze dataset by introducing the Gaze360 dataset, which includes a larger number of subjects and more varied scenarios. They also proposed an architecture that incorporates temporal information, using a CNN backbone to extract features from individual consecutive frames, followed by bidirectional LSTM layers to model temporal information, and an MLP that predicts the gaze direction. Zhang et al. \cite{zhang2017s} proposed the FullFace architecture which is a custom CNN with spatial weights that takes the image as input and outputs the 3D gaze estimate. Chen et al. \cite{chen2018appearance} constructed a model based on dilated convolutions that not only takes as input the face image but also the eye crops to help the model focus more on the ocular region for the final prediction.

Cheng et al. \cite{cheng2020coarse} proposed CA-Net, which uses a coarse-to-fine approach with a CNN that processes the full face image to estimate a basic gaze direction, and another CNN that processes the eye images in order to make a fine prediction. However, the authors constructed a custom CNN model, which is suboptimal as it cannot easily be modified based on new improvements in deep learning. In contrast, our approach enables straightforward updates to every module in the architecture and can utilize any image feature extractors.

The L2CS-Net \cite{abdelrahman2022l2cs} architecture utilizes a ResNet-50 \cite{he2016deep} backbone on the input image and the extracted gaze features are fed to two branches for predicting the yaw and pitch gaze angles. Each branch includes a fully connected layer that generates continuous as well as discrete predictions. Yan et al. \cite{yan2023gaze} improve on this architecture by modifying the backbone to employ strip pooling which makes the receptive field more suited to gaze estimation. They also incorporate multi-criss-cross attention to capture dependencies between the eye features.

Fang et al. \cite{fang2021dual} propose a 3D gaze estimation approach which employs a module for head pose detection and eye features extraction to obtain the final prediction. This information is then utilized in conjunction with a depth map to also detect the gaze target. In contrast, our method does not require any additional labels for head pose or depth as it relies solely on the image input to output the 3D gaze.


\section{Method}
\noindent \textbf{CrossGaze Architecture.}
A high level overview of the proposed architecture is shown in Figure \ref{fig:architecture}. Our model takes as input an in-the-wild image and uses a face detection model to obtain the bounding box predictions of the face and eye regions. The face image is fed into an encoder that captures global gaze features, while the eye images are given as input to the eye encoder which extracts local features. To obtain an eye-informed gaze prediction, the extracted features of the 3 images are combined through a cross-attention module. The resulting combination of features is averaged and processed by a linear layer that outputs the 3D gaze direction.

We consider the face image $I$ of size $H \times W \times 3$ to be the input for the Face Encoder $Enc$ and the images $I', I''$ of size $H' \times W' \times 3$ corresponding to the left and right eyes as the input for the Eye Encoder $Enc'$. The Face Encoder obtains global gaze features $F \in \mathbb{R}^{n \times d}$ ($n$ - number of features, $d$ - feature dimension) that take into account both the visual information around the eye area and outside of it. On the other hand, the Eye Encoder obtains rich local features $F' \in \mathbb{R}^{n \times d}$ as it processes images in which the eyes region has a higher resolution than in the face image. The global and local features are merged by a module which employs cross-attention. The fused features are averaged and processed by an MLP which predicts the 3D gaze. This is formulated as:
\begin{equation}
    \begin{aligned}
    F, F' = Enc(I), Enc'(I', I'') \\
    F'' = CrossAttention(F, F') \\
    \hat{g} =  MLP(AvgPool(F''))\\
    \end{aligned}
\end{equation}    
where $F''$ are the unified features, $AvgPool$ stands for Average Pooling and $MLP$ for Multi-Layer Perceptron. $CrossAttention$ is computed as:
\begin{equation}
    CrossAttention = Softmax(Q'K^{T} / \sqrt{d}) V
    \label{eq:cross-attention}
\end{equation}
where the Queries ($Q' \in \mathbb{R}^{n \times d}$) are obtained from the sequence of local features $F'$ and the sequences of Keys ($K \in \mathbb{R}^{n \times d}$) and Values ($V \in \mathbb{R}^{n \times d}$) are obtained from the sequence of global features $F$.


\noindent \textbf{Implementation Details.}
While many existing methods \cite{abdelrahman2022l2cs,kellnhofer2019gaze360} use polar coordinates as outputs and labels for training, we found that there is no improvement in comparison to the use of 3D vectors. Consequently, we use normalized 3D vectors $(x,y,z)$ as both outputs and labels.





For training the CrossGaze model we utilize the cosine loss which represents the complement of the cosine similarity between the predicted gaze vector $\hat{g}$ and the ground truth $g$:

\begin{equation}
    L_{Cosine} = 1 - {\frac{\hat{g_i} \cdot g_i}{\lVert \hat{g_i} \rVert \cdot \lVert g_i \rVert}}
    \label{eq:cosine}
\end{equation}

We adapt the RandAugment algorithm \cite{cubuk2020randaugment} as training-time augmentation to increase the robustness of the gaze estimation models. For simplicity and to preserve the gaze information, we remove structural augmentations in the form of rotation-based and shear-based transformations. Additionally, we incorporate into the pool of image transformations the cutout augmentation \cite{devries2017improved}, which removes a small patch of the image. Cutout can act as a form of dropout in cases where the removed patch corresponds to the eyes region, further regularizing the gaze estimation model.

The models are trained using the AdamW \cite{loshchilov2017decoupled} optimizer with a batch size of 256 for 200 epochs. The learning rate employs a step schedule that has a starting value of 0.0001, a step size of 10 epochs and a multiplicative factor of decay equal to 0.95. For training the models a single NVIDIA A100 with 40GB of VRAM was utilised. The training of the CrossGaze model takes approximately 10 hours.

\section{Experiments and Results}
We chose to conduct our experiments on the Gaze360 dataset as it is a large-scale dataset containing in-the-wild scenarios. Multiple works \cite{kellnhofer2019gaze360, ghosh2023automatic} demonstrated that models trained on Gaze360 generalize well to other gaze benchmarks, obtaining better results than pretraining on different datasets. Consequently, we conduct all our experiments on the Gaze360 dataset which contains 129K training images and 26K testing images. We utilize 2 subsets of Gaze360: Front 180\degree\ and Front Facing. Front 180\degree\ limits the gaze angles to below 90\degree\ while the Front Facing subset limits them to below 20\degree. This is a similar methodology to other works in gaze estimation \cite{abdelrahman2022l2cs, cheng2021appearance, kellnhofer2019gaze360, fang2021dual}, as the gaze cannot be inferred from images in which the eyes are not visible. Furthermore, all ablation experiments are conducted on the Front 180\degree\ subset which contains approximately 85k training images and 16k testing images.

The most widely used evaluation metric in the field of gaze estimation is the angular error, which is calculated as the angle in degrees between the predicted 3D gaze vector and the ground truth gaze vector. In line with most works on gaze estimation \cite{abdelrahman2022l2cs, kellnhofer2019gaze360, fang2021dual} we also utilize it as the main evaluation metric. The angular error is formulated as:

\begin{equation}
    Angular Error = \arccos{\frac{\hat{g_i} \cdot g_i}{\lVert \hat{g_i} \rVert \cdot \lVert g_i \rVert}}
    \label{eq:angular}
\end{equation} 

\subsection{Evaluation in the Wild}
We present a comparative analysis of our architecture with other state-of-the-art models in appearance-based 3D gaze estimation. \textbf{FullFace} \cite{zhang2017s} employs convolutional modules with spatial weights to predict the 2D and 3D gaze from the input image. \textbf{Dilated-Net} \cite{chen2018appearance} leverages dilated convolutions to capture minor changes in the gaze information. \textbf{RT-Gene} \cite{fischer2018rt} utilizes an ensemble of 4 VGG-16 \cite{simonyan2014very} models on the eye images along with a head pose prediction model to obtain the gaze estimation. \textbf{CA-Net} \cite{cheng2020coarse} utilizes a CNN on the face image to obtain a coarse prediction and separate encoders for the eye images to refine the prediction. \textbf{Gaze360 LSTM} \cite{kellnhofer2019gaze360} uses a CNN on a window of 7 frames and the extracted features are temporally aggregated with a bidirectional LSTM. \textbf{L2CS-Net} \cite{abdelrahman2022l2cs} uses a ResNet-50 as backbone and employs 2 different prediction heads for the yaw and pitch angles. \textbf{SPMCCA-Net} \cite{yan2023gaze} integrates strip pooling and criss-cross attention to the ResNet backbone. \textbf{DAM} \cite{fang2021dual} predicts the head pose and extracts the features of the eye crops to make the gaze prediction, however it benefits from additional annotations.

Table \ref{tab:results-sota} displays the comparison between the proposed CrossGaze model and the other 3D gaze estimation architectures on the Front 180\degree\ and Front Facing subsets of the Gaze360 benchmark. For our model, we show the average and standard deviation of the mean angular error for 3 different runs. The CrossGaze architecture is composed of an Inception ResNet \cite{szegedy2017inception} face encoder, a ResNet-18 \cite{he2016deep} eye encoder, and a cross-attention module that combines the extracted features of both models. The pretrained version of CrossGaze on VggFace2 \cite{cao2018vggface2} achieves state-of-the-art mean angular error on the Front Facing testing subset, demonstrating its capability for gaze estimation in the wild. On Front 180\degree, CrossGaze is competitive with DAM \cite{fang2021dual} which utilizes additional head pose annotations during training. The randomly initialized version of our architecture manages to be on par with the pretrained versions of other models, showcasing its capability to generalize with less data.

\begin{table}[hbt!]
    \caption{Comparison of gaze estimation models on subsets of the Gaze360 dataset. Models highlighted with * employ initialization from pretrained weights while those with ** also utilize head pose annotations. Table adapted from \cite{abdelrahman2022l2cs}.}
    \label{tab:results-sota}
    \begin{center}
     \resizebox{\linewidth}{!}{
        \begin{tabular}{ c | c | c  }
            \textbf{Model} & \textbf{Front 180\degree} & \textbf{Front Facing} \\
            \midrule
            FullFace \cite{zhang2017s} & 14.99\degree & N/A \\
            Dilated-Net \cite{chen2018appearance} & 13.73\degree & N/A \\
            RT-Gene (4 ensemble) \cite{fischer2018rt} & 12.26\degree & N/A \\
            CA-Net \cite{cheng2020coarse} & 11.20\degree & N/A \\
            Gaze360 LSTM* \cite{kellnhofer2019gaze360} & 11.04\degree & N/A \\
            L2CS-Net ($\beta=1$)* \cite{abdelrahman2022l2cs} & 10.41\degree & 9.02\degree \\
            SPMCCA-Net ($\beta=2$)* \cite{yan2023gaze} & 10.13\degree & 8.40\degree \\
            DAM** \cite{fang2021dual} & {9.6}\degree & 9.2\degree \\
            \midrule
            \textbf{CrossGaze (Random Init)} & 10.65\degree \scriptsize{$\pm$ 0.03} & 7.84\degree \scriptsize{$\pm$ 0.27} \\
            \textbf{CrossGaze (Pretrained)} & \textbf{9.94}\degree \scriptsize{$\pm$ 0.06} & \textbf{7.17\degree} \scriptsize{$\pm$ 0.04} \\
        \end{tabular}
        }
    \end{center}
\end{table}

\subsection{Ablation Studies}
In this section, we present ablation studies for each critical component of CrossGaze: the face encoder backbone, pretraining dataset, and methods for incorporating eye information in the computation.  Our CrossGaze architecture is constructed from scratch in a progressive manner, starting from the face encoder backbone and incrementally adding improvements such as the pretraining dataset and the eye feature integration. For all comparisons we show the mean and standard deviation of the performance of each setting computed for 3 different runs.

\begin{table}[hbt!]
    \caption{Performance of each architecture on the Front 180\degree\ subset. The Inception ResNet obtains the lowest error, demonstrating its suitability for gaze estimation in the wild.}
    \label{tab:results-backbones}
    \begin{center}
    \resizebox{\linewidth}{!}{
        \begin{tabular}{c | c | c  }
             \textbf{Model Init.} & \textbf{Model} & \textbf{Mean Angular Error}  \\
            \midrule
             \multirow{4}{*}{Random Init}   
            & EfficientNet & 14.1\degree \scriptsize{$\pm$ 0.32}\\
            & Swin Transformer & 12.03\degree \scriptsize{$\pm$ 0.3}\\
            & ConvNeXt  & 12.79\degree \scriptsize{$\pm$ 0.05}\\
            & Inception ResNet  & \textbf{10.91\degree} \scriptsize{$\pm$ 0.09}\\
            \midrule
             \multirow{4}{*}{Pretrained ImageNet}  
            & EfficientNet &  12.03\degree \scriptsize{$\pm$ 0.3} \\
            & Swin Transformer  & 10.87\degree \scriptsize{$\pm$ 0.05} \\
            & ConvNeXt  & 11.11\degree \scriptsize{$\pm$ 0.08} \\
            & Inception ResNet & \textbf{10.82\degree} \scriptsize{$\pm$ 0.08} \\
        \end{tabular}
    }
    \end{center}
\end{table}

\noindent \textbf{Multi-scale face features aids gaze estimation}. To obtain a strong gaze estimation architecture we experiment with multiple image processing backbones. These models extract the features of the face image which are then passed to a gaze estimation head consisting of an MLP that outputs the 3D vector. We experiment with CNN backbones such as EfficientNet \cite{tan2019efficientnet}, ConvNeXt \cite{liu2022convnet}, Inception ResNet \cite{szegedy2017inception} and an attention-based backbone (i.e. Swin Transformer \cite{liu2021swin}).

The results of this experiment are shown in Table \ref{tab:results-backbones}. In both initialization scenarios, the Inception ResNet architecture obtains the lowest mean angular error with a value of 10.91\degree\ for random initialization and 10.82\degree\ for ImageNet pretrained weights. These results demonstrate the suitability of the Inception ResNet architecture for processing the face: the capability to process the input at multiple scales in every layer helps the model focus both on the eyes region and on the periocular area for the gaze prediction.

\noindent \textbf{Face pretraining enhances gaze feature extraction.}
We analyze the impact of different pretraining datasets on the task of gaze estimation. For this, we only use the top performing face model from the previous experiments, an Inception ResNet, pretrained on the following datasets: ImageNet \cite{deng2009imagenet}, Casia-WebFace \cite{yi2014learning}, and VGGFace2 \cite{cao2018vggface2}. We initialize the Inception ResNet with the pretrained weights of each dataset and train the entire network for gaze estimation.

Table \ref{tab:results-pretraining} shows the results of the experiment, which indicate that the models pretrained on face images obtain better results compared to those pretrained on ImageNet. Face datasets have a closer distribution to the data in the Gaze360 dataset, which naturally translates in improved performance. Additionally, the model pretrained on the larger dataset (VGGFace2) performs better than the model pretrained on Casia-WebFace due to the increased amount of training data. While Casia-WebFace contains approximately 500K images, VGGFace2 contains 3.3M images, resulting in an improvement of 0.24\degree\ in mean angular error. 

\begin{table}[hbt!]
    \caption{Results of pretraining an Inception ResNet on different datasets and transfer learning to gaze estimation.}
    \label{tab:results-pretraining}
    \begin{center}
        \begin{tabular}{ l | c  }
            \textbf{Pretraining Dataset}  & \textbf{Mean Angular Error} \\
            \midrule
            ImageNet \cite{deng2009imagenet} & 10.82\degree \scriptsize{$\pm$ 0.08} \\
            CASIA-WebFace \cite{yi2014learning} & 10.26\degree \scriptsize{$\pm$ 0.05} \\
            VggFace2 \cite{cao2018vggface2} & \textbf{10.02\degree} \scriptsize{$\pm$ 0.07}  \\
        \end{tabular}
    \end{center}
\end{table}

\noindent \textbf{Integrating eye features enriches the gaze information.}
To enable the model to focus on the eye region for the gaze prediction task we insert an additional encoder that separately takes as input both eye images. The extracted face and eye features are combined to obtain the 3D gaze estimation. We experiment with 2 different fusing strategies: the first consists of combining the face and eye features through a fully connected network while the second involves aggregating the information with a cross-attention module.

Table \ref{tab:eye_features} displays the results of the eye features integration for an Inception ResNet face encoder and a ResNet-18 eye encoder. We utilize the ResNet-18 as the secondary encoder because the eye images are of a lower resolution (64x64 after rescaling). In the case of both random initialization and pretrained weights initialization, the use of additional eye features brings an improvement in gaze estimation performance. Furthermore, the cross-attention mechanism manages to better aggregate the information, as it obtains a smaller error in both scenarios compared to the fully connected network. These results motivated the design of the CrossGaze architecture, shown in Figure \ref{fig:architecture}, to also employ the secondary eye encoder and the cross-attention module.

\begin{table}[hbt!]
    \caption{Results for combining features. Cross-attending to eye features improves gaze estimation performance.}
    \label{tab:eye_features}
    \begin{center}
     \resizebox{\linewidth}{!}{
        \begin{tabular}{ c | c | c  }
            \textbf{Model Init.} & \textbf{Eye features combination} & \textbf{Mean Angular Err.} \\
            \midrule
            \multirow{ 3}{*}{Random Init} & No eye features & 10.91\degree \scriptsize{$\pm$ 0.09} \\
            & FCN & 10.75\degree \scriptsize{$\pm$ 0.02} \\
            & Cross-Attention & \textbf{10.65\degree} \scriptsize{$\pm$ 0.03}\\
            \midrule
            \multirow{ 3}{*}{Pretrained VggFace2} & No eye features & 10.02\degree \scriptsize{$\pm$ 0.07} \\
            & FCN & 10.01\degree \scriptsize{$\pm$ 0.05} \\
            & Cross-Attention & \textbf{9.94\degree} \scriptsize{$\pm$ 0.06} \\
        \end{tabular}
        }
    \end{center}
\end{table}

    


\section{Conclusion}
This work proposes CrossGaze, a CNN-based architecture that leverages cross-attention, designed for the task of 3D gaze estimation from images in the wild. On the Gaze360 benchmark, our architecture outperforms several state-of-the-art methods. The model processes both the full face image as well as the eye images to make an eye-informed prediction. To combine the global features of the face with the local images of the eyes, we employ a cross-attention module that captures the most relevant characteristics of the gaze.

We conduct an ablation study, starting from scratch and incrementally adding improvements to the architecture. We start by choosing a suitable face backbone with multi-scale processing at every layer, which helps in extracting relevant gaze features. For this backbone, we show that pretraining on large scale datasets of face images improves generalization, as opposed to other general pretraining datasets. Our results demonstrate that incorporating a separate eye encoder and combining global and local gaze features through cross-attention further improves the performance.   

Our work represents a step forward towards gaze estimation in realistic environments and has the potential to enable practical applications including human-computer interaction, driver assistance systems, and assistive technology for individuals with disabilities.

{\small
\bibliographystyle{ieee}
\bibliography{bibliography}
}

\end{document}